\documentclass{article}

\usepackage{PRIMEarxiv}
\usepackage{listings}
\usepackage[utf8]{inputenc} 
\usepackage[T1]{fontenc}    
\usepackage{hyperref}       
\usepackage{url}            
\usepackage{booktabs}       
\usepackage{amsmath,amssymb,amsfonts}
\usepackage{algorithm}
\usepackage{comment}
\usepackage{amsmath}
\usepackage{algpseudocode}
\usepackage{algorithm}
\usepackage{nicefrac}       
\usepackage{microtype}      
\usepackage{fancyhdr}       
\usepackage{graphicx}       
\graphicspath{{media/}}     
\usepackage{booktabs}
\usepackage{xcolor}
\title{Bridging Logic and Learning: A Neural-Symbolic Approach for Enhanced Reasoning in Neural Models (ASPER)
}

\author{
  Fadi Al Machot\thanks{Corresponding author: Fadi Al Machot (e-mail: fadi.al.machot@nmbu.no)},
\\
  \\
  Faculty of Science and Technology (REALTEK) \\
  Norwegian University of Life Sciences \\ NMBU \\
  1430 Ås, Norway\\
}

\begin{document}
\maketitle

\begin{abstract}
Neural-symbolic learning, an intersection of neural networks and symbolic reasoning, aims to blend neural networks' learning capabilities with symbolic AI's interpretability and reasoning. This paper introduces an approach designed to improve the performance of  neural models in learning reasoning tasks. It achieves this by integrating Answer Set Programming (ASP) solvers and domain-specific expertise, which is an approach that diverges from traditional complex neural-symbolic models. In this paper, a shallow artificial neural network (ANN) is specifically trained to solve Sudoku puzzles with minimal training data. The model has a unique loss function that integrates losses calculated using the ASP solver outputs, effectively enhancing its training efficiency.
Most notably, the model shows a significant improvement in solving Sudoku puzzles using only 12 puzzles for training and testing without hyperparameter tuning. This advancement indicates that the model's enhanced reasoning capabilities have practical applications, extending well beyond Sudoku puzzles to potentially include a variety of other domains. The code can be found on GitHub: https://github.com/Fadi2200/ASPEN.
\end{abstract}

\maketitle

\section{Introduction}
Neural-symbolic learning aims to combine the adaptive learning abilities of neural networks with symbolic AI's clarity and logical capabilities. Thus, research efforts have focused on creating hybrid models that combine Neural Networks with Symbolic Learning, e.g., language translation \cite{bahdanau2014neural}, neural logic machine \cite{dong2019neural}, and logic synthesis from Input-Output examples \cite{chen2017towards}.

This integration represents a step towards addressing some limitations in training neural models, such as the need for large annotated datasets, difficulty handling complex reasoning tasks, and the lack of interpretability in model decisions \cite{MarcusDavis2019}.
While neural-symbolic AI is promising, it faces significant challenges in lacking a generalizable framework that can be easily adapted across different problem spaces. This limitation restricts the broader application of these models and requires extensive customization for new domains \cite{BesoldEtAl2017}.
Addressing these limitations, this work introduces an approach called ASPER that effectively integrates answer set programming (ASP) solvers and domain-specific expertise into neural models.

This paper diverges from traditional neural-symbolic models by focusing on straightforward, effective methods incorporating logical reasoning and expert knowledge. 
The strength of the approach lies in its simplicity and adaptability. The proposed approach does not require extensive datasets. Instead, it leverages the logic in the problem domain to guide the learning process. This aspect is crucial in fields where training data is limited or expensive to collect.
Additionally, the approach enhances the interpretability of the model's decisions. By incorporating rule-based logic into the training process, the model's decisions can be traced back to these underlying rules, providing clearer explanations. This transparency is an advancement over traditional 'black box' models.
Using a custom loss function designed to integrate these logical constraints further represents the unique capabilities of the ASPER. As a result, it ensures that the model not only learns from the available data but also adheres to the logical structure of the problem, resulting in more accurate and reliable outcomes.

\section{Related works}
Recent developments in neural-symbolic AI have revolutionized the integration of low-level data with high-level reasoning capabilities. Central to this advancement are approaches that combine data-driven neural modules with logic-based symbolic systems \cite{HudsonManning2019, dAvilaGarcezEtAl2019, JiangAhn2020, dAvilaGarcezLamb2023}. These combined systems have shown remarkable proficiency in handling complex tasks such as visual question answering (VQA), concept learning, and the development of explainable and revisable models \cite{YiEtAl2018, MaoEtAl2019, CiravegnaEtAl2020, StammerEtAl2021}.

Deep Probabilistic Programming Languages (DPPLs) \cite{ManhaeveEtAl2018} is a probabilistic logic programming language that seamlessly integrates deep learning through neural predicates. This pioneering framework uniquely combines general-purpose neural networks with sophisticated probabilistic-logical reasoning, capable of both symbolic and subsymbolic inference. Furthermore, NeurASP \cite{YangEtAl2020}  is an innovative extension of answer set programs that integrates neural networks for enhanced computation. It treats neural network outputs as probability distributions for atomic facts in answer set programs, offering a straightforward method to merge sub-symbolic and symbolic processes. Moreover, Scallop \cite{HuangEtAl2021}  is a system that combines deep learning with symbolic reasoning in a scalable way, addressing limitations in DPPLs.

The previous approaches have introduced innovative ways to integrate neural and symbolic modules, offering a scalable approach to AI systems. These systems leverage neural predicates to enhance probabilistic reasoning,  making deep learning systems more flexible and efficient.
Despite these advancements, scalability remains a significant challenge in DPPLs.
The primary issue lies in the increasing computational complexity and demand for substantial memory and processing power. Moreover, scaling up risks overfitting. Additionally, as these systems grow in complexity, maintaining user-friendly interfaces that allow efficient model building, training, and deployment without requiring deep expertise in neural and symbolic AI domains is crucial for their practical scalability. 

In addition,  recent studies have continued to explore various neural network architectures to tackle the challenge of solving Sudoku puzzles. For instance, a Neuro-Symbolic approach has been proposed, which blends deep reinforcement learning techniques with symbolic learning, aiming to improve systematicity and generalization across larger sets of inputs. This method emphasizes the use of logic predicates to approach the problem systematically. It has shown a promising success rate when solving the puzzles within a defined number of steps \cite{neurosymbolic2020sudoku}. However, the primary limitation noted is the consideration of only 3, 5, and 8 empty cells in their experiments.

Another innovative approach is the development of SATNet, a differentiable maximum satisfiability solver incorporating CNNs. SATNet operates using a fast coordinate descent approach to solve associated semidefinite programs, showing potential in the field of deep learning for solving systematic problems like Sudoku \cite{wang2019satnet}. However, the researchers utilized a substantial dataset for training, with their experiments based on 9,000 training examples and 1,000 test examples.

This work (ASPER) simplifies the integration of domain expert knowledge into neural models.
The empirical evaluations demonstrate the effectiveness of the proposed approach in reducing the complexity of potential solutions and facilitating end-to-end training of complex data structures by employing a model that learns from both data and embedded logical constraints. This makes the proposed method more scalable and adaptable to various problem domains where reasoning and logic are essential.

\section{Answer Set Programming  (ASP)}

Answer set programming (ASP), as outlined in \cite{gelfond1988stable} and \cite{niemela1999logic}, has gained prominence in the field of artificial intelligence (AI) due to its robustness in knowledge representation and reasoning. Notably recognized for its expressiveness, ASP excels in handling scenarios with incomplete information. An ASP program typically comprises two segments: a knowledge base with factual information and a set of rules delineating the problem-solving approach. The solutions in ASP are formulated as answer sets or models, essentially possible solutions that comply with the given constraints.

In the language of AnsProlog, also known as A-Prolog, an ASP program is constructed with rules like:
\begin{equation}
    a_0 \leftarrow a_1, \ldots, a_m, \neg a_{m+1}, \ldots, \neg a_n,
\end{equation}
where \( 1 \leq m \leq n \). Each \( a_i \) denotes an atom in the propositional language, with \( \neg a_i \) signifying a negation-as-failure literal (naf-literal). The rule is structured such that the head (left-hand side) and the body (right-hand side) fulfil distinct roles. Notably, a rule can exist with an empty head, termed a constraint, or with an empty body, known as a fact, but cannot have both simultaneously absent.

Consider \( X \) as a set of ground atoms in a given ASP program, forming its Herbrand base. The body of a rule in the format of (1) is satisfied by \( X \) if the intersection of \( \{a_{m+1}, \ldots, a_n\} \) and \( X \) is empty, and \( \{a1, \ldots, a_m\} \) is a subset of \( X \). A rule with a non-empty head satisfies \( X \) if either \( a0 \) is in \( X \) or its body is not satisfied by \( X \). A constraint is fulfilled by \( X \) if its body is unsatisfied by \( X \).

\section{Methodology}
The proposed approach, designed for adaptability and generalization across a range of problem domains, seamlessly integrates three key components: the Answer Set Programming (ASP) problem solver, a neural model, and a customized loss function. The ASP solver is crucial for translating complex domain knowledge into structured logic, making it adaptable to diverse scenarios ranging from for example, puzzles, physics, and biology to strategic business problems. The neural model, trained on empirical data and structured knowledge from the ASP solver, is uniquely equipped for adaptability in various environments, particularly where training data is limited. The core innovation lies in the design of the loss function, which comprises overall data-driven learning loss, constraints loss derived from ASP solver solutions for logical consistency, and domain expertise constraints for aligning with expert knowledge. This integrated loss function ensures a balanced learning process, combining data, logical constraints, and expert insights.

\subsection{Problem Definition}
Board games, particularly puzzles, typically involve a grid-based structure where each cell can contain an element from a predefined set. The objective often revolves around arranging these elements according to specific rules. Formally, such a grid can be represented as a matrix \( G \) where \( G_{i,j} \) represents the element in the \( i^{th} \) row and \( j^{th} \) column. The problem can be defined as finding a matrix \( G \) that satisfies a set of conditions which are unique to each game but generally involve constraints on the arrangement of elements. These conditions might include:

\begin{enumerate}
    \item Each cell \( G_{i,j} \) must contain an element from a given set, e.g., \( G_{i,j} \in \{1, 2, ..., N\} \) for a game with \( N \) different elements.
    \item Certain elements may need to be aligned or arranged in specific patterns, rows, columns, or subgrids, depending on the game's rules.
    \item Additional constraints unique to the specific puzzle or game, dictating how elements can be placed or moved within the grid.
\end{enumerate}

The challenge in these games often lies in the logical arrangement or sequence of moves required to achieve a final, 'solved' state of the grid, adhering to all specified rules and constraints.

\subsection{Combined Loss}
For any given puzzle, the puzzle should be represented in ASP. This involves encoding the puzzle elements and constraints as logical rules. The general form of these rules can be expressed as:
 
\begin{equation}
\text{asp\_rule(features, constraints, expert\_insights)}
\end{equation}

These rules ensure the ASP solver considers all possible configurations that satisfy the puzzle's constraints.

\subsubsection{Standard loss}

Parallel to the ASP formulation, a neural model should be employed. This model uses a customized loss function that combines a standard loss with a puzzle-specific constraints-based loss. The standard loss ($L_{standard}$) can be generally defined using a squared error
(MSE) or  cross entropy as:

\begin{equation}
L_{\text{standard}} = -\sum_{i=1}^{C} y_i \log(p_i)
\end{equation}

This equation, $L_{standard}$ represents the Cross-Entropy loss for a multiclass classification problem. Here, $C$ is the total number of classes, $y_i$ is the binary indicator (0 or 1) if class label $i$ is the correct classification for the observation, and $p_i$ is the predicted probability of the observation being class $i$.

\subsubsection{Constrained loss}
The constraints-based loss $L_{constraints}$ for a generic puzzle is formulated to penalize solutions that violate the puzzle's specific rules and constraints. It can be generalized as:

\begin{equation}
L_{\text{constraints}} = \sum_{\text{all constraints}} \text{penalty}(constraint_{violation})
\end{equation}

This loss ensures the model's predictions adhere to the puzzle's logical structure.

\subsubsection{Domain Expert Knowledge loss}

Integrating domain expert knowledge, such as specialized strategies, significantly enhances the interpretability and accuracy of machine learning models. Answer Set Programming (ASP) solvers are key to this integration. The results obtained from ASP solvers are particularly beneficial as they provide structured, logical outcomes that can be leveraged for improved learning.

\begin{equation}
L_{\text{expert}} = \sum_{\text{all expert rules}} \text{penalty}(\text{deviation from expert rule})
\end{equation}

\subsubsection{ASPER Approach - Combined Loss Function and Model Training}
The combined loss function ($L_{combined}$) integrates both standard and constraints\-based losses:

\begin{equation}
L_{\text{combined}} = \alpha L_{\text{standard}} + \beta L_{\text{constraints}} + \gamma L_{\text{expert}}
\label{combined}
\end{equation}

Here, $\alpha$ and $\beta$ are weighting factors that balance the influence of standard accuracy and adherence to constraints. The model is trained on data generated by solving puzzles with ASP, to minimize $L_{combined}$. This ensures that the model learns both the accuracy in puzzle element prediction and adherence to the puzzle's intrinsic rules.

\begin{algorithm}
\caption{ASPER - A General Methodology for Neural Learning with ASP Solver Integration}
\begin{algorithmic}[1]
\Require Data, DomainKnowledge, Expertise
\Ensure Trained Model
\Function{TrainDeepLearningModel}{Data, DomainKnowledge, Expertise}
    \State ASP $\gets$ Initialize ASP solver with domainKnowledge
    \State Model $\gets$ Initializethe Neural Model
    \State $\alpha, \beta, \gamma \gets$ Initialize weights for loss components
    \State LossFunction $\gets$ Create Custom Loss Function with Expertise Knowledge
    \For{epoch $\gets$ 1 to Epochs}
        \ForAll{puzzle in Data}
            \State StructuredData $\gets$ ASP.solve(puzzle)
            \State Prediction $\gets$ Model.predict(puzzle)
            \State $L_{\text{standard}} \gets$ CalculateStandardLoss(Prediction, StructuredData)
            \State $L_{\text{constraints}} \gets$ CalculateConstraintsLoss(Prediction, StructuredData)
            \State $L_{\text{expert}} \gets$ CalculateExpertLoss(Prediction, StructuredData)
            \State $L_{\text{combined}} \gets \alpha L_{\text{standard}} + \beta L_{\text{constraints}} + \gamma L_{\text{expert}}$
            \State Model.update\_weights($L_{\text{combined}}$)
        \EndFor
    \EndFor
    \State \Return Model
\EndFunction
\Function{SolvePuzzleWithModel}{Puzzle, Model}
    \State Prediction $\gets$ Model.predict(Puzzle)
    \State Solution $\gets$ Post-process Prediction to satisfy game rules
    \State \Return Solution
\EndFunction
\end{algorithmic}
\end{algorithm}

Algorithms 1 shows the pseudocode details a methodical approach for integrating neural models with Answer Set Programming (ASP), highlighting a multi-faceted training process. It begins by initializing an ASP solver with domain knowledge, ensuring that the model is attuned to the specific constraints and logic of the problem domain. Consequently, the neural model is set up to learn from empirical data. The idea lies in the custom loss function, which combines standard data-driven learning loss, constraints-based loss from the ASP solutions, and a domain expertise loss. This combination ensures that the learning is data-centric, logically coherent, and aligned with expert insights. The training loop iterates over the data, utilizing the ASP solver for structured problem-solving and the deep learning model for prediction, thus striking a balance between empirical learning and rule-based reasoning. While demonstrated in the context of puzzles, this methodology has a broad scope of applicability across various domains.

\section{Use-Case - Suduku}
In the realm of Sudoku, a puzzle is defined as a 9x9 grid where each cell may contain a digit from 1 to 9. The challenge lies in filling the grid so that each row, column, and 3x3 subgrid contains all digits from 1 to 9 exactly once. Formally, the Sudoku grid can be represented as a matrix \( S \) where \( S_{i,j} \) represents the value in the \( i^{th} \) row and \( j^{th} \) column. The problem can be defined as finding a matrix \( S \) that satisfies the following conditions for all \( i, j \) in the range 1 to 9:

\begin{enumerate}
    \item \( S_{i,j} \in \{1, 2, ..., 9\} \) for all \( i, j \).
    \item Each number appears exactly once in each row: \( \sum_{j=1}^{9} [S_{i,j} = k] = 1 \) for each \( k \) and each \( i \).
    \item Each number appears exactly once in each column: \( \sum_{i=1}^{9} [S_{i,j} = k] = 1 \) for each \( k \) and each \( j \).
    \item Each number appears exactly once in each 3x3 subgrid.
\end{enumerate}

In the proposed ASP-based approach for solving Sudoku, the puzzle's constraints are encoded as rules in an ASP program. This program uses the ASP solver `clingo` to find solutions satisfying these constraints. The key aspects of this implementation, based on the provided code, are (see Listing 1):

\begin{lstlisting}[language=Prolog, caption={ASP code for Sudoku constraints}, label=lst:asp_code]
1  { cell(Row,Col,Val) : Val=1..9 } = 1 :- Row=1..9, Col=1..9.
2  :- cell(Row,Col1,Val), cell(Row,Col2,Val), Col1 != Col2.
3  :- cell(Row1,Col,Val), cell(Row2,Col,Val), Row1 != Row2.
4  :- cell(Row1,Col1,Val), cell(Row2,Col2,Val),
5     Row1 != Row2, Col1 != Col2, (Row1-1)/3 = (Row2-1)/3, 
(Col1-1)/3 = (Col2-1)/3.
\end{lstlisting}

\begin{enumerate}
    \item \textbf{Line 1} defines a rule that ensures each cell in the grid has exactly one number between 1 and 9.
    \item \textbf{Line 2} enforces that no two cells in the same row can have the same value.
    \item \textbf{Line 3} ensures that no two cells in the same column have the same value.
    \item \textbf{Line 4-5} imposes the constraint that no two cells in the same 3x3 subgrid have the same value.
\end{enumerate}

By leveraging ASP, we can efficiently solve Sudoku puzzles, regardless of complexity. This declarative approach allows for a concise and clear puzzle representation, facilitating intuitive and straightforward solution processes.

\subsection{Loss functions of 
Suduku}
ASPER  integrates Answer Set Programming (ASP) with a deep learning model to solve the defined Sudoku problem. In ASP, we express the constraints of Sudoku as rules. For instance, the row constraint is encoded as:

\begin{equation}
\sum_{j=1}^{9} \text{cell}(i, j, v) = 1 \quad \forall i, v
\end{equation}

Similar rules are constructed for columns and 3x3 subgrids, ensuring that the ASP solver finds solutions that respect the Sudoku constraints.

The deep learning model, tailored to solve Sudoku, uses a loss function that combines the standard cross-entropy with a constraints-based loss. The cross-entropy, representing the standard loss ($L_{standard}$), is defined as:

\begin{equation}
L_{\text{cross-entropy}} = -\frac{1}{81} \sum_{i=1}^{9} \sum_{j=1}^{9} S_{i,j} \log(\hat{S_{i,j}})
\end{equation}

where $S_{i,j}$ represents the predicted value for the cell in the $i_{th}$  row and $j_{th}$ column.

The constraints-based loss ($L_{constraints}$) is formulated to penalize solutions that violate Sudoku's row, column, and subgrid rules. It can be expressed as:

{
\scriptsize
\begin{equation}
\begin{split}
L_{\text{constraints}} = \sum_{num=1}^{9} \Biggl( 
& \sum_{i=1}^{9} \left( 1 - \sum_{j=1}^{9} \text{is\_empty}_{i,j} \times [y_{i,j} = num] \right)^2 \\
& + \sum_{j=1}^{9} \left( 1 - \sum_{i=1}^{9} \text{is\_empty}_{i,j} \times [y_{i,j} = num] \right)^2 \\
& + \sum_{sub=1}^{9} \left( 1 - \sum_{(i,j) \in \text{sub}} \text{is\_empty}_{i,j} \times [y_{i,j} = num] \right)^2 \Biggr)
\end{split}
\end{equation}
}

This loss function accounts for the frequency of each number in each row, column, and subgrid, focusing on empty cells in the original puzzle.

Assuming that the domain knowledge expert loss calculates the difference in the sum of predicted numbers and actual numbers in rows, columns, and subgrids. It ensures that the total sum of numbers in each row, column, and subgrid in the prediction matches that in the true solution. The loss is defined as:

\begin{equation}
\begin{split}
L_{\text{expert}} = \sum_{i=1}^{9} \Biggl( 
& \left| \sum_{j=1}^{9} y_{\text{pred\_row}_i, j} - \sum_{j=1}^{9} y_{\text{true\_row}_i, j} \right| \\
& + \left| \sum_{j=1}^{9} y_{\text{pred\_col}_i, j} - \sum_{j=1}^{9} y_{\text{true\_col}_i, j} \right| \\
& + \sum_{j=1}^{9} \biggl| \sum_{(k, l) \in \text{subgrid}_{i,j}} y_{\text{pred\_subgrid}, k, l} \\
& \quad - \sum_{(k, l) \in \text{subgrid}_{i,j}} y_{\text{true\_subgrid}, k, l} \biggr| \Biggr)
\end{split}
\end{equation}

While the $L_{constraints}$ loss ensures each number appears exactly once per row, column, and subgrid, the $L_{expert}$ specifically ensures that the sum of the numbers in each row, column, and subgrid in the predicted solution matches the sum in the true solution.

The constraints-based loss uses a squared difference to penalize deviations from Sudoku rules. In contrast, $L_{expert}$ uses the absolute difference between the sums of predicted and true values, which could be more sensitive to deviations in number distribution.
The $L_{expert}$ loss is particularly effective in scenarios where maintaining the sum of numbers is crucial, reflecting a deeper Sudoku-solving strategy often employed by experienced players.

Combining these two losses with the neural model standard loss functions in a neural network model for solving Sudoku puzzles enables the model to learn the basic rules of Sudoku and incorporate more advanced solving strategies, enhancing its problem-solving capabilities.

The combined loss function ($L_{combined}$) is a weighted sum of the standard loss and the constraints loss:

\begin{equation}
L_{\text{combined}} = \alpha L_{\text{standard}} + \beta L_{\text{constraints}} + \gamma L_{\text{expert}}
\label{combinedAll}
\end{equation}

Training the model involves generating puzzles, solving them with ASP to create training data, and then using this data to train the neural network to minimize $L_{combined}$. This process ensures that the model not only learns to predict the numbers in each cell accurately but also respects the fundamental rules of Sudoku, thus effectively solving the puzzle. 

\section{Implementation and Evaluation}
The Sudoku solving process employs a shallow neural network constructed using Keras' Sequential architecture \cite{chollet2015keras}. This model starts with a Dense layer of 64 neurons and ReLU activation, processing a flattened 81-element array representing the 9x9 Sudoku grid. It then includes a Dense layer (9x9x9) for outputting potential cell values, reshaped into a 9x9x9 tensor and finalized with a softmax activation layer. Optimization uses the Adam optimizer and TensorFlow's GradientTape for gradient computation \cite{abadi2016tensorflow}. Training involves custom loss functions, including cross-entropy and Sudoku-specific constraints, to ensure both accuracy and adherence to Sudoku rules. This method is supported by Clingo \cite{gebser2019multi}, an ASP solver that verifies the correctness of Sudoku puzzles generated for training, using KFold cross-validation to enhance model robustness and unbiased accuracy.

The evaluation of accuracy is conducted through a KFold cross-validation process. Each fold's data is split into training and validation sets, and the model is trained using custom loss functions. Post-training, the model's predictions on the validation set are compared to actual solutions. Accuracy is calculated based on the proportion of correctly predicted cells across all Sudoku puzzles in the test set. These accuracy metrics from each fold are then aggregated to comprehensively assess the model's overall performance in solving the puzzles.

\section{Results}
\begin{table*}[h]
\centering
\begin{tabular}{|p{2cm}|p{1.5cm}|p{2cm}|p{2cm}|p{2cm}|p{2cm}|}
\hline
\textbf{\# Puzzles} & \textbf{difficulty} & \textbf{Standard Only} & \textbf{Standard + Expert} & \textbf{Standard + Constraints} & \textbf{All Combined} \\ \hline
12 & 0.1 & 0.84  & 0.86  & 0.92  & 0.88 \\ \hline
12 & 0.3 & 0.70  & 0.69  & 0.74 & 0.70 \\ \hline
12 & 0.6 & 0.48 & 0.41 & 0.49  & 0.51  \\ \hline
12 & 0.8 & 0.26 & 0.28  & 0.28 & 0.29\\ \hline
100 & 0.1 & 1.00 & 1.00  & 1.00 & 1.00 \\ \hline
100 & 0.3 & 1.00 & 1.00 & 1.00 & 1.00 \\ \hline
100 & 0.6 & 0.19  & 0.20 & 0.20  & 0.20  \\ \hline
1000 & 0.8 & 0.11  & 0.11  & 0.11  & 0.11 \\ \hline
\end{tabular}
\caption{Comparative Evaluation of Various Loss Combinations on 12, 100, and 1000 Sudoku Puzzles Across Difficulty Levels 0.1, 0.3, 0.6, and 0.8. The data presented reflects outcomes from a 3-fold cross-validation process, with the ± symbol indicating the standard deviation observed across different folds. This comprehensive analysis provides insights into the effectiveness of each loss combination under varying puzzle complexities and sizes}
\label{tab:results}
\end{table*}

\begin{figure*}[h]
\centering
\includegraphics[width=0.9\textwidth]{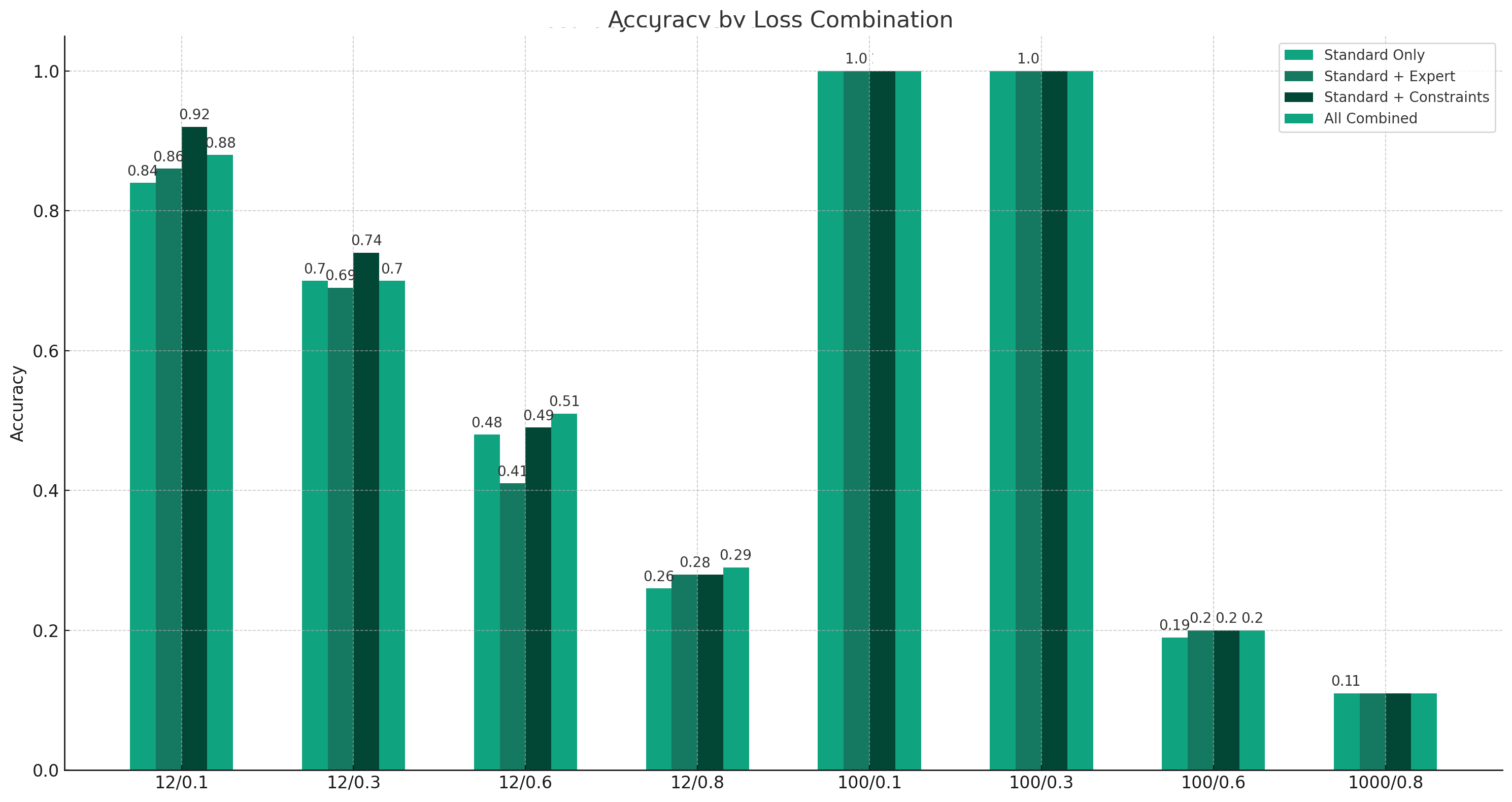}
\caption{Evaluation of Various Loss Combinations on 12, 100, and 1000 Sudoku Puzzles Across Difficulty Levels 0.1, 0.3, 0.6, and 0.8. The data presented reflects outcomes from a 3-fold cross-validation process.}
\label{fig:plot}
\end{figure*}

The results detailed in Table \ref{tab:results} show the critical importance of selecting an optimal loss function specifically tailored to the distinct characteristics of a problem, such as the number of puzzles and the difficulty level, particularly in the context of Sudoku puzzles. This study underscores the power of employing a hybrid loss approach to tackle complex challenges. The exploration involves various loss functions in training a deep learning model for solving Sudoku puzzles, including the different loss combinations. The analysis indicates that with higher difficulty levels in Sudoku puzzles, which typically means less available information, there is a noticeable decrease in accuracy across most loss combinations. This trend highlights deep learning models' challenges in effectively utilizing limited information. The analysis further shows that the smaller the dataset used for training, the more is the impact of ASPER on the overall performance. In addition, ASPER shows that neural models can handle complex, less predictable puzzles with limited training data under constrained training conditions.
 The All Combined approach shows high performance, particularly in scenarios with lower difficulty levels and smaller puzzle sets, suggesting its adeptness in handling complex scenarios. Interestingly, in larger sets (100 puzzles), differences between loss functions become less pronounced, while in smaller sets (12 puzzles), the differences are significant, with the Combined loss showing its strength. The study highlights the importance of constraint learning in deep learning, which is crucial for model development in domains where rules and constraints play a pivotal role. It also stresses the integration of domain expertise, which enhances the model's problem-solving capabilities, a principle extendable to other fields for augmented AI performance. The role of cross-validation in this context ensures that the model's performance is unbiased, thereby bolstering the reliability of the results. Importantly, the approach's potential extends beyond Sudoku, with applicability to other complex tasks such as natural language processing and image recognition.
 
\begin{figure*}[h]
\centering
\includegraphics[width=0.8\textwidth]{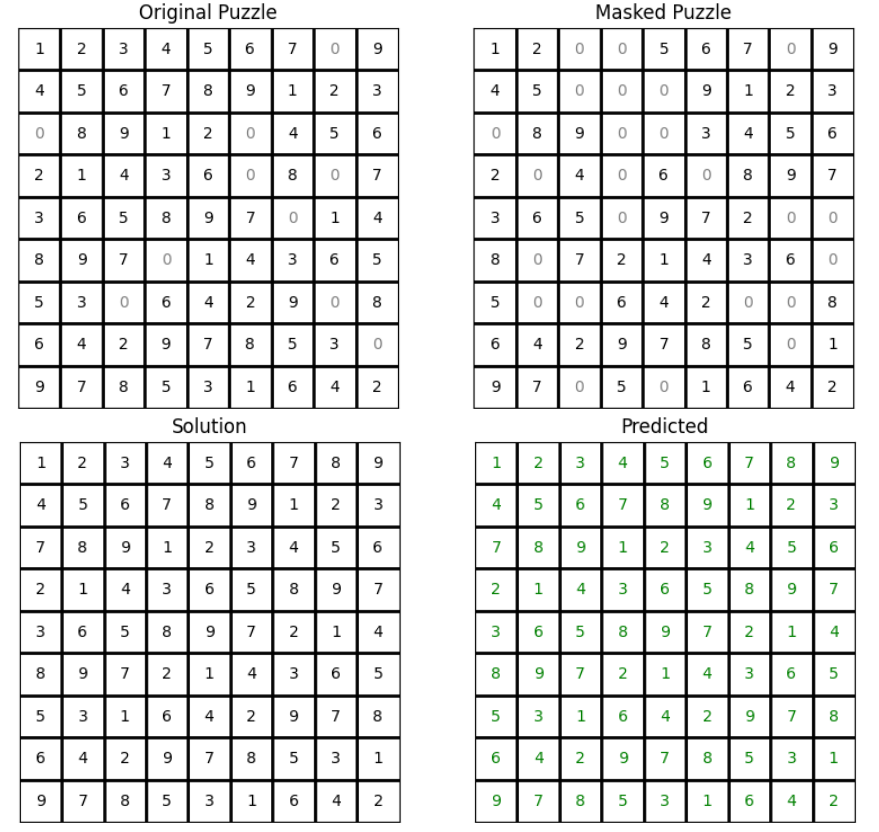}
\caption{The image displays a set of four 9x9 Sudoku grids, each representing a different stage of puzzle solving. The top left grid is the "Original Puzzle" with all cells filled, indicating the starting complete Sudoku. The top right grid is the "Masked Puzzle," showing the Sudoku with certain numbers omitted, which presents the challenge to be solved. The bottom left grid is the "Solution," providing the completed puzzle with all numbers correctly placed. The bottom right grid is the "Predicted" solution, presumably produced by a Sudoku-solving algorithm, with correct predictions highlighted in green and incorrect ones in red. This visualization serves to compare the model's performance against the actual solution.}
\label{fig:puzzles}
\end{figure*}
\subsection{Discussion}
ASPER reveals key insights into using different loss functions in neural-symbolic AI, particularly in solving Sudoku puzzles. The finding shows that using standard loss ($L_{standard}$) resulted in low accuracy when the number of training data is small. This outcome underscores the insufficiency of using a standard loss-based approach for complex problems like Sudoku.
In contrast, the incorporation of constraints-based loss ($L_{constraint}$) and domain expert knowledge loss ($L_{exopert}$) led to an improvement when the number of training data is small, considering different difficulty levels. This highlights the  role of integrating domain-specific knowledge and Sudoku-specific rules into the learning process, illustrating the value of tailored approaches in solving combinatorial problems. ASPER  shows improved performance across varied data samples and complexity levels, indicative of its adaptability to diverse real-world scenarios. By using different datasets for each set of conditions, we ensure the model's effectiveness is not an artefact of specific data characteristics, thereby reducing the risk of overfitting and enhancing the reliability of ASPER findings. However, as the volume of training data increases, the incremental benefits of ASPER become less noticeable, aligning with expectations that extensive data can inherently enhance deep learning model performance.

ASPER contributes to the evolving field of neuro-symbolic AI by showcasing the successful integration of domain expertise into deep learning models. This approach is particularly effective in complex reasoning tasks and scenarios with limited data, combining neural networks' pattern recognition capabilities with symbolic AI's structured, logical reasoning. This method reduces the reliance on extensive datasets and enhances the model's deductive reasoning abilities.
Moreover, ASPER addresses the crucial issue of AI interpretability and transparency. Embedding logical rules into training provides more traceable and understandable decision-making pathways. This is especially important in fields where understanding the reasoning behind AI decisions is critical.
The methodology of embedding logical rules directly into the training process can enhance the decision-making transparency in many AI applications, a critical factor in fields such as healthcare, finance, and commercial systems where understanding the rationale behind AI decisions is as vital as the decisions themselves.

\begin{table}[h!]
\centering
\begin{tabular}{|c|c|c|c|}
\hline
\textbf{Method} & \textbf{Input} & \textbf{Number of Data for Training} & \textbf{Accuracy of Solution} \\ \hline
(2018) \cite{park2018sudoku} & Textual Sudoku (9x9) & 1 Million & 70.0\% \\ \hline
(2018) \cite{palm2018recurrent} &  Textual 
Sudoku (9x9) & 216,000 & 96.6\% \\ \hline
(2019) \cite{wang2019satnet}&  Textual Sudoku (9x9) & 9000 & 98.3\% \\ \hline
(2023) \cite{neurosymbolic2020sudoku}&  Textual Sudoku (9x9) & - & 100\% (8 empty cells only) \\ \hline
\end{tabular}
\caption{Comparison of different methods for solving Sudoku.}
\label{table:sudoku_methods}
\end{table}

Table  \ref{table:sudoku_methods}) shows that ASPER  outperforms the benchmarks state-of-the-art benchmarks in terms of handling different difficulty levels and achieving a competitive accuracy a with a minimal dataset.
It is also worth highlighting that integrating expert knowledge and constraints into the learning process has improved the accuracy of puzzles with difficulty levels of 0.1 and 0.3. This integration appears to improve the model performance and provide the ability to leverage additional information effectively, which is not solely reliant on the data-driven patterns learned from the input data.

\subsection{Conclusion}
This paper proposes the ASPER approach to explore the integration of Answer Set Programming (ASP) with neural models. The findings suggest that incorporating ASP's logical reasoning into the framework of neural models can enhance their performance, particularly in combinatorial problems. This integration appears to offer a  way to embed domain-specific knowledge and rules into AI models, an area of growing interest in machine learning.
A key observation from this paper is the potential improvement in model accuracy when incorporating constraints-based loss. The ASPER approach demonstrates that a combined loss methodology, which integrates various aspects of learning and reasoning, may offer a balanced path forward in the development of more sophisticated AI systems.
Furthermore, the work addresses aspects of AI interpretability and transparency by embedding logical rules directly into the training process to enhance the decision-making transparency of AI models. This approach could be particularly beneficial in sectors where understanding the rationale behind AI decisions is required.
Looking ahead, the ASPER approach could be scaled and adapted to various applications, though this would require careful consideration of each domain's unique requirements and constraints. The paper contributes to the ongoing discourse in the AI community about balancing data-driven insights with rule-based logic, particularly in the context of ethical AI development and deployment.

\bibliographystyle{plain}
\bibliography{neuro}
\end{document}